\documentclass[runningheads]{llncs}
\usepackage{graphicx}
\let\llncssubparagraph\subparagraph
\let\subparagraph\paragraph
\usepackage[compact]{titlesec}
\let\subparagraph\llncssubparagraph
\usepackage[shortlabels]{enumitem}
\usepackage{url}
\usepackage{booktabs}
\usepackage{adjustbox}
\usepackage{array}
\usepackage{multirow}
\usepackage{siunitx,booktabs}
\usepackage{capt-of}% or \usepackage{caption}
\usepackage{mathtools} % Bonus
\usepackage{todonotes}
\usepackage{amssymb}
\usepackage{mathtools}
\usepackage{dsfont}
\usepackage{algorithm}
\usepackage{algpseudocode}
\usepackage{makecell}
\usepackage{subfig}
\newcolumntype{R}[2]{%
    >{\adjustbox{angle=#1,lap=\width-(#2)}\bgroup}%
    l%
    <{\egroup}%
}

\renewcommand{\epsilon}{\varepsilon}

\newcommand{\sobol}{Sobol\kern-0.15em\'{ } }

\newcommand*\rot{\rotatebox{45}}

\usepackage{etoolbox}

\BeforeBeginEnvironment{figure}{\vskip-2ex}
\AfterEndEnvironment{figure}{\vskip-1ex}

\begin{document}
\title{Evaluating the Robustness of Deep-Learning Algorithm-Selection Models by Evolving Adversarial Instances}

\titlerunning{Evaluating the Robustness of Deep-Learning Algorithm-Selection Models}

\author{Emma Hart\orcidID{0000-0002-5405-4413} \and Quentin Renau\orcidID{0000-0002-2487-981X} \and
Kevin Sim\orcidID{0000-0001-6555-7721} \and Mohamad Alissa\orcidID{0000-0002-9548-863X}}
\authorrunning{Hart et. al.}
\institute{Edinburgh Napier University\\
\email{\{e.hart, q.renau, k.sim, m.alissa\}@napier.ac.uk}}

\maketitle              % typeset the header of the contribution
\begin{abstract}
Deep neural networks (DNN) are increasingly being used to perform algorithm-selection in combinatorial optimisation domains, particularly as they accommodate input representations which avoid designing and calculating features. Mounting evidence from domains that use images as input shows that deep \textit{convolutional} networks are vulnerable to adversarial samples, in which a small perturbation of an instance can cause the DNN to misclassify. However, it remains unknown as to whether \textit{deep recurrent networks (DRN)} which have recently been shown promise as algorithm-selectors in the bin-packing domain
are equally vulnerable. We use an evolutionary algorithm (EA) to find perturbations of instances from two existing benchmarks for online bin packing  that cause trained DRNs to misclassify: adversarial samples are successfully generated from up to  $56\%$ of the original instances depending on the dataset.

Analysis of the new misclassified instances sheds light on the `fragility' of some training instances, i.e. instances where it is trivial to find a small perturbation
that results in a  misclassification and the factors that influence this.
Finally, the method generates a large number of new instances misclassified with a wide variation in confidence, providing a rich new source of training data to create more robust models.

\end{abstract}

\section{Introduction}
\label{sec:intro}

For most combinatorial optimisation domains, it is well known that for a given set of solvers, each can perform differently on different instances, with no single algorithm dominating the others. This gives rise to the need to perform per-instance algorithm selection, described in detail in a survey by Kerschke {\em et. al.}~\cite{kerschke2019automated}.

Typically a machine-learning algorithm is trained to predict the best solver for an instance. While earlier works generally relied on training models using feature-vectors derived from instances, more recent works have exploited new deep-learning models which circumvent the need to derive features.  Deep \textit{convolutional} neural network architectures originally developed for image-classification have been shown to be capable of learning to predict the best solver from a portfolio in the TSP domain by representing the instance as an image containing the locations of each city~\cite{seiler2020deep,prager2022automated}.  In other work, deep \textit{recurrent} neural networks such as LSTM (long short-term memory)~\cite{hochreiter1997long} and GRU (gated recurrent network)~\cite{cho2014learning} which take an ordered sequence of tokens as input have been trained  to act as an algorithm-selectors in the bin-packing ~\cite{alissa2019algorithm,alissa2023automated} and vehicle routing~\cite{diaz2023addressing} domains.

However,  mounting evidence from the machine-learning literature shows in particular that deep \textit{convolutional} networks that use images as input  are vulnerable to \textit{adversarial attacks}: that is, applying 
a small perturbation $\delta$ to an instance $x$ leads to the perturbed instance $x+\delta$ being  misclassified ~\cite{alzantot2019genattack,su2019one,qiu2021black}.

As a result, an increasing amount of research  is being directed towards generating adversarial attacks, with the goal of understanding how robust a classifier is to variations in input samples. Most of this research considers \textit{black-box} attacks~\cite{narodytska2017simple}: a scenario in which the attacker only has access to the inputs and outputs of a model, and has no information about the architecture or weights of the model itself.

Attacks can be \textit{targeted} in that the adversarial sample is optimised to output a specific (incorrect) class, or\textit{ untargeted}, in which case the goal is simply to maximize the loss between the predicted class and the true class. However,  to the best of our knowledge, there has been no attempt to understand the extent to which deep \textit{recurrent} networks which have been proposed  as algorithm-selectors in combinatorial settings such as bin-packing and VRP (vehicle routing problem) are robust to perturbations in input data. This is crucial to understand: in any real-world setting  it is reasonable to assume that a selector may be faced with many instances whose data are very similar to those that the model was trained on (i.e. are \textit{in-distribution}) and therefore will be classified correctly, but a systematic methodology to investigate the extent to which this is true is lacking.

To address this issue we propose a method to determine the extent to which DRNs trained as algorithm-selectors in the bin-packing domain are vulnerable to small changes in input data. We select online bin-packing as a domain for two reasons:  (1) the class of problems are NP-hard and are worth studying because they appear as a factor in many other kinds of optimization problem~\cite{ross2003learning}; (2) DRN models have been shown to be highly accurate classifiers in combinatorial settings~\cite{alissa2023automated,diaz2023addressing}. We propose an evolutionary algorithm (EA) to evolve a mask that when applied to an instance from a dataset used to train a model  causes a small modification that results in the perturbed instance being misclassified.
The contributions are as follows:

\begin{itemize}
\item Defining an EA to evolve adversarial instances of bin-packing instances  that are misclassified by a trained DRN while ensuring that the modified instance remains in-distribution.
\item Providing new evidence that trained DRNs are vulnerable to adversarial attacks using two datasets and associated models.
\item Providing new insights into which instances are particularly susceptible to attack and those that are robust, shedding light on which instances lie  close to the decision-boundaries of the classifiers
\item  Generation of a  very large set of new instances which can be used for future training: these instances are diverse in terms of the output probability of the correct class, i.e. the confidence with which an instance is classified.
\end{itemize}

\section{Related Work}

Deep convolutional neural-networks have rapidly garnered interest within combinatorial optimisation for the purpose of algorithm-selection~\cite{kerschke2019automated}. Loreggio \textit{et. al.}~\cite{loreggia2016deep} convert a textual description of instances from the SAT and CSP domains\footnote{SAT: Satisfiability, CSP:  Constraint Satisfaction Problems~\cite{Hoos24}} to an image and train a convolutional neural network to output solver predictions.  Seiler \textit{et. al.}~\cite{seiler2020deep} also use a convolutional NN trained on images derived from TSP instances to predict the best solver.   In continuous optimisation, Prager \textit{et. al.} ~\cite{prager2022automated}  consider both image and point cloud representations of the COCO benchmarks and train a type of CNN called \textit{shuffleNET}~\cite{ma2018shufflenet} to select the best performing algorithm. 

Instead of using image-based input, Alissa \textit{et. al.}~\cite{alissa2019algorithm,alissa2023automated} directly use the textual description of an instance in  the bin-packing domain (i.e. an ordered list of item-sizes) to train two types of deep recurrent neural-networks (DRNs) to predict the best solver. They compare two types of DRN  --- LSTM~\cite{hochreiter1997long} and GRU~\cite{cho2014learning} --- to feature-based classifiers, finding that the GRU  achieves within 5\% of the oracle performance on between 80.88 and 97.63\% of the instances, depending on the dataset. Diaz~\cite{diaz2023addressing} propose an attention-based transformer network for selecting a solver for in the VRP domain, finding improved performance compared to a multi-layer perceptron.

However, despite the increasing focus in the use of deep classifiers in algorithm-selection in the combinatorial domain, we are unaware of any work which has investigated the extent to which these classifiers are robust to perturbations in instances via generating adversarial samples.
Liu {\em et. al,.}~\cite{liu2023promoting} propose a method to promote data diversity for learning-based branching modules in branch-and-bound (B\&B) solvers in which a  learning-based solver and instance augmentation policy are adversarially trained, but do  not focus per-se  on the robustness of the model, rather on instance generation. In contrast, several recent studies in the domain of image-classification have used evolutionary algorithms to generate adversarial samples that represent attacks against models trained on well-known datasets to illustrate the vulnerability of a trained model~\cite{su2019one,alzantot2019genattack} (often described as an `attack' on the network). For example, \textit{GenAttack}~\cite{alzantot2019genattack} evolves visually imperceptible adversarial examples against state-of-the-art image recognition models trained on three popular datasets (ImageNet, CIFAR-10 and MNIST) with orders of magnitude fewer queries than previous approaches in a black-box setting. Given an input image, it creates a population by applying random modifications to the pixels of the original image. 
In~\cite{qiu2021black}, the authors  compare three evolution strategies to generate  black-box adversarial attacks on networks trained on the ImageNet dataset, evolving reduced dimensionality samples that are then scaled up to reduce the computational burden. Their results show that CMA-ES~\cite{HansenO01} is particularly effective in finding adversarial samples with the fewest queries. In contrast to the works just described which modify multiple or even all pixels in an image, in~\cite{su2019one}, differential evolution (DE)~\cite{StornP97} is used to evolve a modification to a \textit{single pixel} showing that current DNNs used for image classification are vulnerable to very low-dimensional attacks. Lin {\em et. al.}~\cite{lin2020black} propose a technique called Black-box Momentum Iterative Fast Gradient Sign Method (BMI-FGSM) that is also inspired by differential evolution, as well as by iterative gradient-based methods.
It leverages DE to approximate gradient direction by searching for the gradient-sign, generating adversarial samples that are hard to detect and which successfully attack DNNs trained on MNIST, CIFAR10, and ImageNet datasets.

Inspired by successful attempts to illustrate the vulnerability of convolutional networks by using an EA to generate attacks, we develop a methodology to evaluate the robustness of deep recurrent network classifiers that use a sequence of tokens as input, using bin-packing as a case-study.  Unlike the work in image-classification which  tends to treat the task as a continuous optimisation problem, we evolve a mask consisting of discrete values in the range $-n \le x \le n$ which indicates how each of the discrete variables describing a combinatorial optimisation instance should be modified.

\section{Methods}

We assume a target model $\mathcal{M}$ that has been trained to output the probability of selecting a target solver $s$ for a given domain. $\mathcal{M}$  can only be queried as a black-box function, i.e. the inputs and outputs are known but there is no information about the model itself.
We search for a perturbation of an \textit{original} instance $i_O$ (from the training data of the  model) labelled as  won by the $k$th solver $s_k$  from a portfolio
such that the \textit{perturbed} instance $i_p$ is now misclassified by the model  $\mathcal{M}$.  Specifically, we consider without loss of generality a scenario in which $k=2$, i.e. there are two available solvers. For each of the \textit{original} instances, there is therefore a winning solver $s_w$ and a losing solver $s_l$, determined according to a metric that quantifies the quality of the packing produced by the solver. For each instance, applying a perturbation can result in two types of misclassification:
\begin{enumerate}
\item The perturbed instance is won by the same solver $s_w$ as the original instance, but the model outputs $s_l$.
\item The perturbed instance is now won by $s_l$, but the model still outputs $s_w$.
\end{enumerate}

We seek to maximise the probability associated with the output of the \textit{incorrect} class. If $p_w$ is the probability output by the network for the winning solver, and $p_l = (1-p_w)$ the probability of the losing solver, then we maximize  $o=p_l-p_w$.  Positive values of $o$ indicate the instance is misclassified while negative values  indicate a correct classification with the magnitude of $|o|$ indicating the classifier confidence in the prediction\footnote{the approach can be generalised to $n$ classes by taking the difference between  $argmax(p_l)$ and $p_w$}.

\subsection{Online Bin-Packing}
\label{sec:bp}
We use the general method described above to evolve adversarial instances in the online bin-packing domain. An instance consists of a sequence of items which must be packed strictly in the order they arrive and no information about the sequence is known in advance of each item arriving. The goal is typically to minimise the number of bins. Simple packing heuristics that determine which bin each item should be placed in are surprisingly effective, particularly best-fit (BF) and first-fit (FF)~\cite{garey1981approximation}. BF places each item into the feasible bin that minimises the residual space, while FF places each item into the first feasible bin that will accommodate it.
The quality of the resulting packing $O_{Falk}$ is defined by the commonly used Falkenaeur metric~\cite{falkenauer1996hybrid} which returns a value between $0$ and $1$ where $1$ is optimal, and rewards packings that minimise empty space. For a given portfolio of solvers, the \textit{winning} solver is defined as the solver that maximises  $O_{Falk}$.
Note that two instances that have identical \textit{sets} of item-sizes but differ in the order in which items arrive can result in different packings, and elicit different performances from each heuristic. Therefore, the ordering of items is an important characteristic of an instance, in addition to the distribution of item-sizes.  In order to minimise the size of the perturbation applied to an instance, we only evolve perturbations that result in a small modification to the size of each item defining an instance. Evolving perturbations that changed the item ordering would result in very different instances and therefore defeat the objective of trying to understand whether small changes to an instance can result in a misclassification.

\subsection{Data and Models}
We use the data and models previously described in \cite{alissa2023automated}. Two datasets denoted (DS2, DS4) are used: each dataset consists of $2{,}000$ instances, of which $50\%$ are solved best by the best-fit (BF) heuristic and the remaining $50\%$ by the first-fit (FF) heuristic. Item sizes are generated from a normal distribution in the range (20,100)
for each dataset; DS2 has 120 items and DS4  has 250. Bins have a maximum capacity of 150. As we previously showed that a Gated Recurrent Network (GRU) outperforms an LSTM on these datasets~\cite{alissa2023automated}, we restrict the current study to evaluating the vulnerability of GRU models only.  We use the GRU architecture and parameterisation as described in~\cite{alissa2023automated} except for the final layer which is replaced with a softmax function~\cite{Softmax} in order to predict probabilities rather than classes.
Models are trained using DS2 and DS4 respectively, according to a $10$-fold cross-validation procedure.
For DS2, the model has a mean accuracy  of $94.44\%$ $(+/- 2.92\%)$   on a validation set (over $10$ folds) and $93.5\%$ on the test set of $400$ instances. For DS4, it achieves $97.00\%$ $(+/- 1.11\%)$ mean accuracy on a validation set and $95.5\%$ on a balanced test set of $400$ instances.

In a preliminary step, we apply the trained models to the full dataset of $2{,}000$ instances, then remove the small minority of instances that are misclassified, given that we are interested in generating perturbed variations of instances which result in an instance originally classified correctly now being misclassified. This results in a small reduction in the size of the datasets used from here on in, by 16 instances for DS2 and 46 instances for DS4.

\subsection{Algorithm Details}
\label{sec:algorithm}
\textit{Individual Representation}:  We evolve a \textit{mask} that is used to perturb an individual instance. The mask contains $i$ integer values, where $i$ is equal to the number of items in the instance. Each integer in the mask can take one of three values $[-1,0,1]$:   the item at position $j$ in the original instance is modified by adding the integer from the mask at position $j$, hence decreasing the item size by 1, doing nothing, or increasing the item size by 1. The modification is restricted to this range in order to minimise the extent of the total change that can be made though clearly could be adapted to allow a larger magnitude of perturbation. The maximum \textit{total change} to the item sizes for an instance is thus equal to the number of items. Note that the modification is not cumulative, i.e. a mask is applied exactly once to the \textit{original  instance}.

\textit{Algorithm}: We use a generational EA to evolve masks.
An initial population of size $P$ is initialised as follows: first, every element in each mask is set to 0. Then, with probability $p_{init}$, each element is uniformly randomly changed to $[-1, 0, 1]$. 
Following the evaluation of the initial population, a generational loop first selects $n=P$ parents; crossover and mutation are applied with probability $p_c, p_m$ respectively to produce an offspring population, after which individuals are evaluated. The offspring population entirely replaces the parent population. Tournament selection is used with tournament size 2 and one-point crossover. Mutation works as follows: each element of the mask is mutated with probability 1/(number of items). A customised mutation operator selects a new value $[-1,0,1]$ with equal probability.  If a modification results in an item size which falls outside of the fixed range of item sizes defining a dataset, then the value is clipped to the respective minimum/maximum allowed value.  The
population size is set to $50$, and the algorithm runs for $500$ generations. Tuning was deliberately kept to a minimum in order to demonstrate that a `default' algorithm is capable of finding adversarial samples.

\textit{Evaluation Function}: As noted above, we consider a setting with two solvers. 
After a mask is applied to an instance $i$, then a packing is produced from each solver, resulting in objective values $o_{BF}, o_{FF}$ according to the Falkenauer metric.
Assume we label the \textit{winning} solver $s_w$ and the \textit{losing} solver $s_l$, and that the probabilities output by the classifier for $s_w, s_l$ respectively are $p_w$ and $p_l$ (such that $p_w +  p_l = 1$), then the fitness function that drives the search for misclassified instances is defined as:
\begin{equation}
f = p_l - p_w
\end{equation}

The function aims to maximise the confidence of an incorrect classification.
A positive fitness value corresponds to a misclassified instance and a negative value to a correctly classified instance. A fitness of exactly $0$ implies both probabilities are equal to $0.5$ however in practice this situation is never encountered.

\begin{algorithm}
\caption{Pseudo-code for assigning  a fitness score to a perturbation}\label{alg:fitness}
\begin{algorithmic}
\Require Mask, $s_1$, $s_2$ \Comment s1,s2 = solver 1, solver 2 
\State $i_P \gets perturb\_instance(i, mask)$
\State $o(s_1) \gets solve\_instance(i_P,s_1)$ \Comment{Falkenauer  metric of $s_1$ on perturbed instance
}
\State $o(s_2) \gets solve\_instance(i_P,s_2)$ \Comment{Falkenauer metric of $s_2$ on perturbed instance
}
\State $winner \gets argmax(o(s_1),o(s_2))$ \Comment winner is id of solver with max fitness
\State $p(s_1) \gets query\_classifier(i_P)$
\State $p(s_2) \gets (1-p(s_1))$
\If{winner is $s_1$}   
    \State $fitness \gets p(s_2)-p(s_1)$
\ElsIf{ winner is $s_1$}
    \State $fitness \gets p(s_1)-p(s_2)$
\EndIf
\State $return(fitness)$
\end{algorithmic}
\end{algorithm}

\subsection{Experimental Protocol}
We conduct an initial experiment in which  we randomly sample $500$ masks with $p_{init}$ set to the relatively high probability of $0.3$ following the initialisation procedure outlined in Section~\ref{sec:algorithm} and apply the masks to each instance. The purpose of this is to gain some insight into how easy it is to generate an adversarial sample by simply randomly sampling masks. 
An instance is labelled as \textit{fragile} if at least one of the randomly sampled masks results in an instance that is misclassified. For DS2, $875 = 43.4\%$ of instances are not labelled fragile, while for DS4, only $51$ instances $= 2.55\%$ are not fragile. This result immediately indicates the model trained on the DS4 instances is much less robust than the DS2 model: randomly sampling masks that on average have $30\%$ of the elements set to $+1$ or $-1$ results in a misclassified instance. In the remaining experiments, all fragile instances are removed from the datasets and we only attempt to evolve masks to perturb the remaining non-fragile instances.  All experiments are repeated $10$ times per instance.

\section{Results}
We first present data from experiments that evaluate the effectiveness of the approach in terms of number of instances that produced  misclassifications and the computational effort required, followed by a deeper analysis of the results.

\subsection{Effectiveness of the EA}
We consider $p_{init}$ values of $\{0.05, 0.3\}$ representing populations that are initialised with very few item perturbations (at $p_{init}=0.05$) and one with approximately $1/3$ of the items being perturbed on average. We measure the \textit{success-rate}, defined as the percentage of the original instances for which the EA evolved at least one mask that resulted in a misclassification across the $10$ runs. To measure the effort required to find an adversarial sample, we define \textit{queries} \footnote{Following the terminology employed in the literature on evolving adversarial samples for image-classification}
as the median of the minimum number of evaluations across each of $10$ runs needed to find an adversarial sample.
Finally, for each instance, we record the type of misclassification as a \% of the successful instances:   T1 -  the true labels of all new misclassified instances are the same as the original instance; T2 - the true labels of all new misclassified instances  are different from the original instance; T3 - the new misclassified instances are from both T1 and T2. 

Table~\ref{tab:summaryData} shows the data just described. For DS2, the success rate decreases as $p_{init}$ decreases, as would be expected: for the same number of evaluations, the evolved masks are likely to contain fewer modifications. Adversarial  samples are found for approximately $33\%$ of instances.
Although the success-rate is higher for DS4 ($\approx 56\%$), recall that the number of non-fragile instances to which the EA is applied is very small ($51$ vs $875$ for DS2). Interestingly, the value of $p_{init}$ has no effect on DS4 ---  in fact, the set of instances for which a successful perturbation is found is identical for both values of $p_{init}$. This suggests there is a subset of instances that are relatively easily perturbed (i.e. at $p_{init} = 0.05$) but further perturbations have no effect. We shed more insight into this in Section \ref{sec:analysis}.

\subsection{Quality of  evolved adversarial instances}

Table~\ref{tab:summaryData} shows the median, the first, and third quartiles of the maximum fitness obtained per instance across the $10$ repeated runs. 
Recall that positive values indicate that a misclassified instance was found. 
For DS2(0.3), a large interquartile difference of positive values ranging from just greater than $0$ (weak confidence in the misclassification) to $1$ (very strong confidence in the misclassification) is observed. 
For DS2(0.05), some masks are evolved that reduce the probability of a correct classification (compared to the original instance) but are not misclassified ($f<0$). 
In contrast for DS4, evolved masks generally create instances that are strongly misclassified. 

\begin{table}[ht]
    \centering
        \caption{Effectiveness of EA: T1,T2,T3 are the types of misclassification as a \% of misclassified instances. Median, first quartile (Q1), and third quartile (Q3) of the maximum fitness obtained at the end of each run over all instances.}
\begin{tabular}{cccccc|ccc}
\hline
\multicolumn{1}{l}{} & Success Rate (\%) & Queries & T1 & T2 & T3 & \begin{tabular}[c]{@{}c@{}}Median\\ fitness\end{tabular} & \begin{tabular}[c]{@{}c@{}}Q1\\ fitness\end{tabular} & \begin{tabular}[c]{@{}c@{}}Q3\\ fitness\end{tabular} \\ \hline
DS2(0.3)             & 33.49             & 1500    & 27 & 61 & 12 &  0.9912  &  0.0002   &  0.9999 \\
DS2(0.05)            & 28.69             & 5600    & 39 & 54 & 7  &  0.9542 & -0.993    &  0.9999 \\
DS4(0.3)             & 56.87             & 50      & 34 & 45 & 21 & 0.9999 & 0.9999   & 1.0   \\
DS4(0.05)            & 56.87             & 2300    & 41 & 41 & 18 & 0.9999 &  0.9999 & 0.9999   \\ 
\hline
\end{tabular}
    \label{tab:summaryData}
\end{table}

\section{Analysis}
\label{sec:analysis}

\subsection{Path towards misclassification}

\begin{figure*}
    \centering
    \subfloat[]{\includegraphics[width=0.32\textwidth]{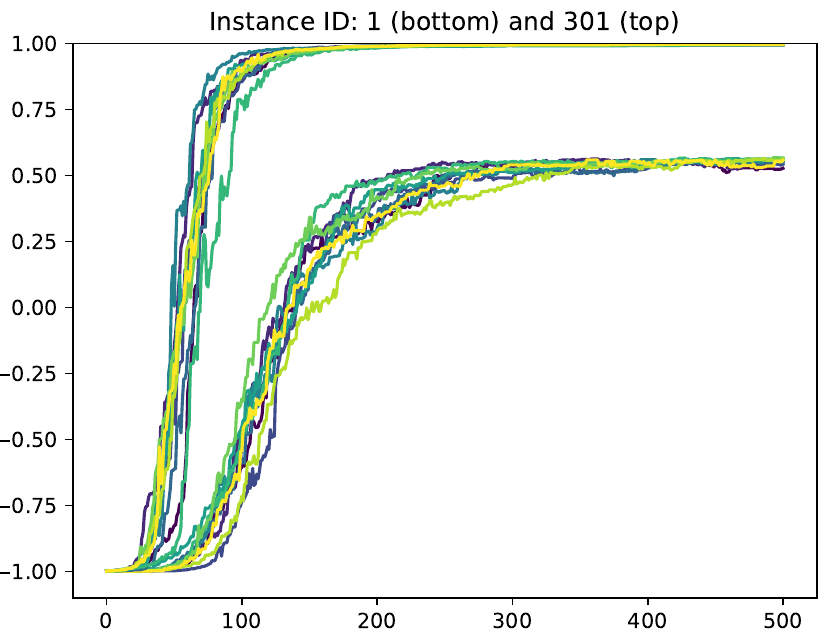}} 
    \subfloat[]{\includegraphics[width=0.32\textwidth]{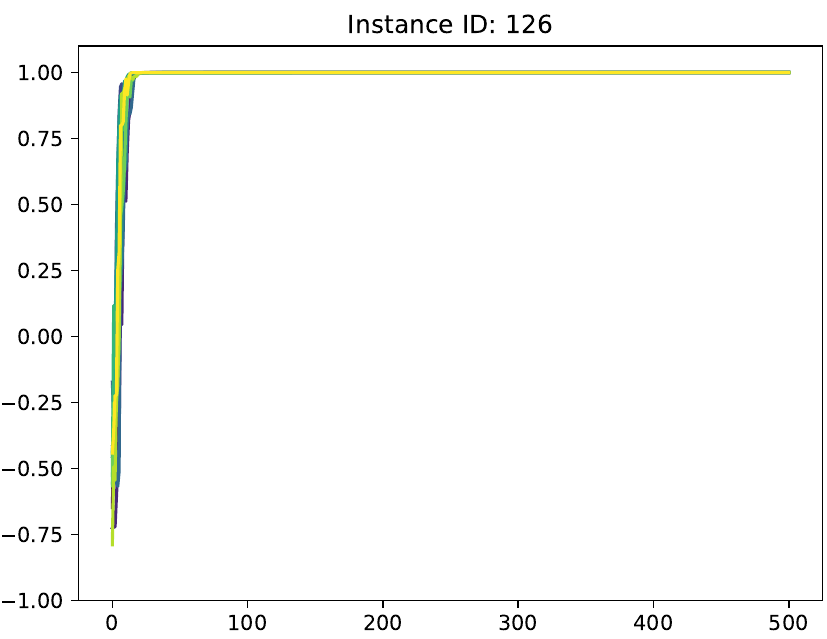}}
    \subfloat[]{\includegraphics[width=0.32\textwidth]{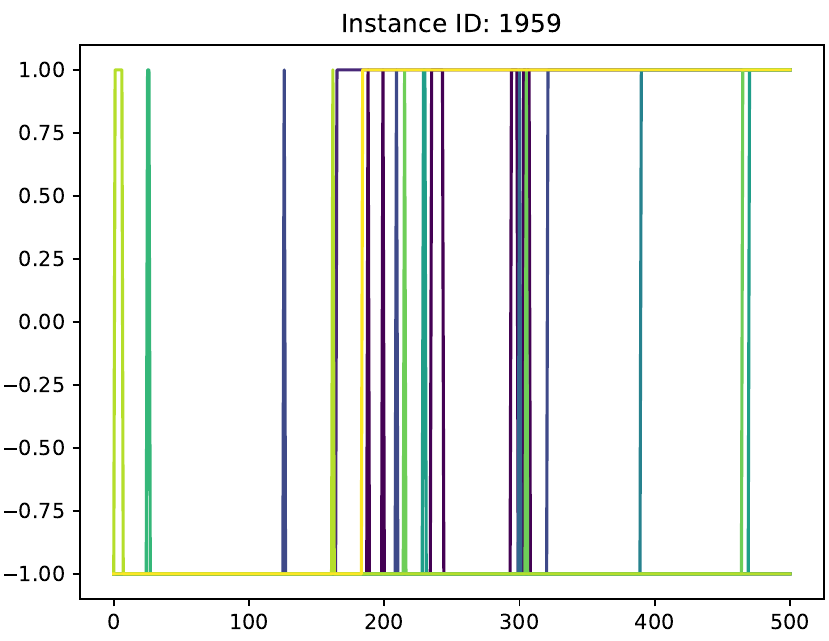}}
    \caption{Behaviours observed during mask evolution: generations (x-axis) vs fitness (y-axis). Each color represents one run. (a) gradual increase in fitness overtime leading to instances classified incorrectly with medium confidence/high confidence; (b) a mask is found in the first few generations which immediately `flips' the classification to strongly misclassified (c)  like (b) but perturbations result in oscillation between high confidence (correct class) and high confidence (incorrect class).} 
    \label{fig:exampleRuns}
\end{figure*}

Recall that each instance is perturbed by the application of a single mask and that fitness is defined as $|p_l-p_w|$, with positive values indicating a misclassification.  Positive values close to $0$ indicate low confidence in the classification, and values close to $1$ indicate very high confidence in the classification. Figure~\ref{fig:exampleRuns} shows some examples of types of fitness curves obtained from evolving masks that successfully cause a misclassification on three instances from DS2. It is clear that  different types of  behaviours are observed. For example, in Figure~\ref{fig:exampleRuns}(a) there is a gradual improvement  of fitness over time, terminating in either a new instance misclassified with medium confidence or high confidence respectively. 
In Figure~\ref{fig:exampleRuns}(b),  a mask is found in the first few generations which causes an immediate `flip' from the original instance being classified correctly with very high confidence to classified incorrectly with very high confidence. In the final example shown (c), successive generations result in an oscillation between high confidence (correct class) and high confidence (incorrect class). This implies that the original instance appears very susceptible to perturbation.

\subsection{Insights into how the instances change}
The EA is restricted to evolving masks that can only modify the size of each item by $(+/-)1$. Given that item sizes in the original datasets are discrete values drawn from a uniform distribution between $20$ and $100$ (and clipped to the minimum/maximum values) then even if every item is changed by $+1$ or $-1$, the distribution of item-sizes is unlikely to deviate from the original normal distribution\footnote{We evaluated this hypothesis empirically by sampling a large number of pairs of (original-instance, modified-instance); a Kolmogorov-Smirnov test showed that the null hypothesis could never be rejected.}.

\begin{figure}
    \centering
    \subfloat[DS2]{\includegraphics[width=0.45\textwidth]{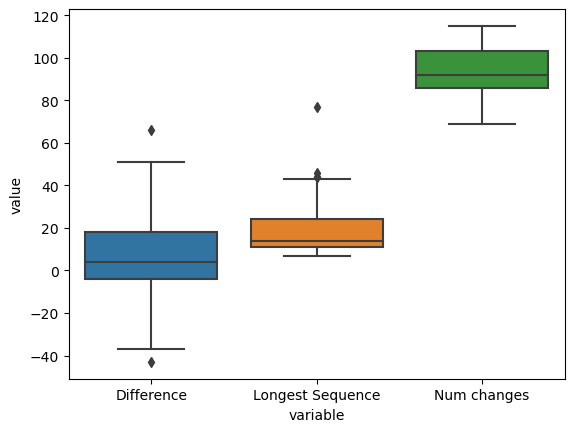}}
    \subfloat[DS4]{\includegraphics[width=0.45\textwidth]{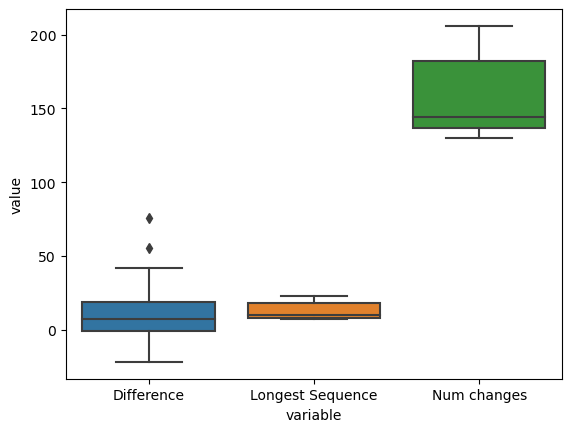}} 
    \caption{Statistics calculated over \textit{all} the new misclassified instances generated from DS2 and DS4, $p_{init}=0.3$. \textit{Difference}: difference in the sum of item sizes between the original instance and a modified instance; \textit{Longest Sequence}: maximum length of a consecutive perturbation; \textit{Changes}: total number of modifications per instances.}
    
    \label{fig:newInstanceStates}
\end{figure}

Given that the bin size remains fixed at $150$, then the sum of the $n$ item sizes $\Sigma=\sum_{i=1}^{i=n} itemSize_i$ in the instance can influence the number of bins required. In the extreme case for example, if all item sizes are increased by $1$, then more bins might be required.  For all of the original instances where it was possible to evolve a mask that resulted in a misclassified  instance, we calculate the difference $D$ between the sum $\Sigma$ of the items in an original instance  and every new misclassified instance produced for that instance. A box plot of the median value of $D$ over all $m$ misclassified instances produced is shown in Figure \ref{fig:newInstanceStates} for DS2. The median difference is $+4$.
Figure~\ref{fig:newInstanceStates} also plots the median number of changes induced by a mask, i.e. the  number of items that will be modified. The median is $92$. This indicates that although a large proportion of the items change size by $+/- 1$ ($\approx 77\%$ of items), the overall effect of these changes is to more or less cancel each other out in terms of $\Sigma$. Therefore it is reasonable to assume that this has little impact on the optimal number of bins required.

Both heuristics pack items strictly in the order specified by the instance. Therefore if consecutive items are modified,  this could potentially alter the packing, thereby causing a misclassification. Figure~\ref{fig:newInstanceStates} also shows the median length of the longest consecutive sequence of modifications in a mask that results in a 
misclassification, where a modification is defined as either an increase or decrease in the item-size. The median length for DS2 is $14.25$, and $10$ for DS4: the sequences are relatively short compared to the instance length of 120/250 respectively. 
The median length of sequences with only positive ($+1$) modifications which might be expected to have more effect on packing is $5$ for both datasets.

\begin{table}
     \caption{Spearman correlation coefficient between the fitness of misclassified instances and statistics describing the change in instance properties.}
      \label{tab:correlations}
    \centering
    \begin{tabular}{clll}
        \hline
       Dataset  & Statistic & Correlation & p-value \\
         \hline
       DS2 & Longest sequence  & -0.898 & $<<0.001$  \\
         & Number of changes  & -0.903 & $<<0.001$  \\
        & Difference & -0.451 & $<<0.001$  \\
       DS4 & Longest sequence  & -0.886& $<<0.001$  \\
       & Number of changes  &- 0.834 & $<<0.001$  \\
       & Difference & -0.623 & $<<0.001$  \\
       \hline
    \end{tabular}
\end{table}

To determine if there is a relationship  between the statistics depicted in Figure~\ref{fig:newInstanceStates} and the median fitness of the misclassified new instances obtained from each of the original instances, we calculate the Spearman correlation coefficient. The results are shown in Table~\ref{tab:correlations}. A very strong negative correlation is obtained both  between fitness and the longest sequence, and fitness and number of changes for DS2 and DS4 ($<-0.8)$; in DS4 there is also a strong correlation ($<-0.6$ between the difference in the sum of item-sizes $D$ and fitness. This suggests that modifying a contiguous sequence does have an influence on the packing, potentially modifying the winning heuristic and resulting in a misclassification.

\begin{figure*}
    \centering
    \subfloat[Instance 565, median fitness of new instances $\approx 0.007$]
    {\includegraphics[width=\textwidth]{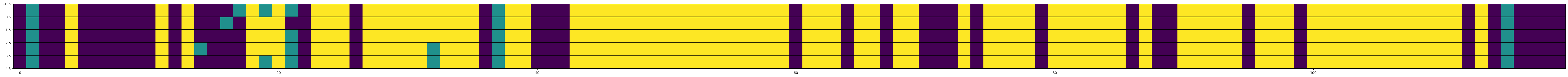}}\\
    \subfloat[Instance 694, median fitness of new instances $0.474$]
    {\includegraphics[width=\textwidth]{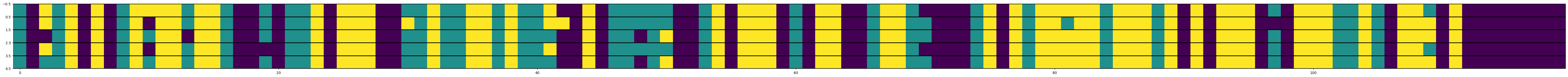}}\\
     \subfloat[Instance 72,  median fitness of new instances $0.999$]
    {\includegraphics[width=\textwidth]{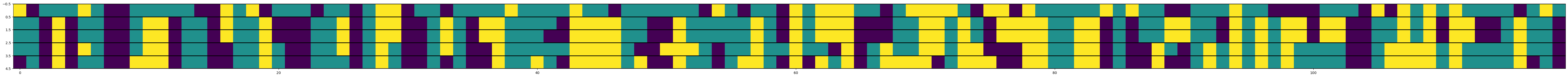}} 
    \caption{DS2: Illustrative examples of masks leading to misclassified instances. Yellow indicates a change of +1, teal 0 and purple -1. $5$ misclassified instances are shown for each original instance.}
    %565,694 ,72
    \label{fig:exampleNewInstances}
\end{figure*}

In Figure~\ref{fig:exampleNewInstances} we visualise examples of masks leading to misclassifications for three example instances: (a) the median fitness  of the new misclassified instances generated from the single original instance is close to $0$, i.e. new instances are misclassified  with very low confidence; (b) the median fitness is $\approx 0.5$, i.e. there is medium confidence in the (mis)classification; (c) the median fitness $\approx 1$, i.e.  very high confidence. This figure clearly illustrates different patterns in the evolved masks. For new instances that are misclassified but with very low confidence ($p$ only just greater than $0$) there are repeated regions where there are contiguous modifications of $+1$, as well as repeated but shorter contiguous regions of '-1' modifications. Very few elements are not modified. In contrast, when the probability of the (mis)classification is close to $1.0$, we only observe very short sequences of each `type' of modification, and it is clear there are many more elements that are not modified at all.

\begin{figure}
    \centering
    \includegraphics[width=0.55\linewidth]{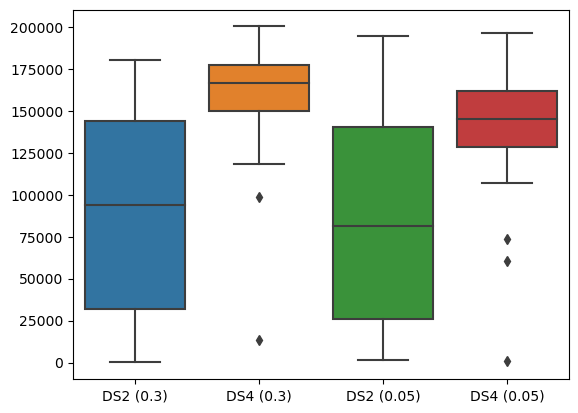}
    \caption{New misclassified instances generated from each of the original instances that produce at least one misclassification (aggregated over $10$ runs).
    }
    \label{fig:numNew}
\end{figure}

Figure~\ref{fig:exampleNewInstances} show only a few of many thousands of masks generated during runs of the algorithm. The EA tries to maximise the difference in output probabilities $p_l-p_w$.  As demonstrated in Figure~\ref{fig:exampleRuns}, in many runs there is a smooth increase in fitness over generations: at each generation,  any individual with fitness $>0$ is misclassified such that $0\le p_l-p_w  \le w$. Therefore, many misclassified instances are discovered during the search process which guides the EA to maximise the fitness function. We count the number of \textit{unique} masks $m_U$ that produced a fitness $>0$ for each starting instance $i$. Figure~\ref{fig:numNew} shows the distribution of $m_U$ over the $a$ instances for which we were able to evolve at least one mask causing a misclassification for each of the two datasets. Statistics are provided in Table~\ref{tab:numNew}. The total number of adversarial samples generated is vast, providing a rich source of training data for training new models. Notice that the median number of adversarial samples discovered \textit{per instance} is much higher for DS4 than DS2, indicating that the model is much  less robust but  that these samples come from very few instances ($29$). On the other hand, as previously noted, there are more instances in DS2 that can be modified ($293$) but the number of adversarial samples generated per instance is lower than DS2.

\hskip-0.5cm

\begin{table}
\parbox{.45\linewidth}{
\centering
               \captionof{table}{Unique misclassified instances. }  \label{tab:numNew}  
                    \begin{tabular}{cccc}
                   
    \hline
        &  \thead{Modifiable\\ Instances}   & \thead{Total \\Instances} & \thead{Median per\\ modifiable \\instance
        }\\
    \hline
        DS2  & 293 &25,810,846 & 93,834\\
        DS4 &  29 & 4,578,680 & 166,633\\
        DS2  & 251 & 21,194,879 & 81,607\\
        DS4  & 29 & 4,003,832& 144,862\\
    \hline
    \end{tabular}

}
\hfill
\parbox{.45\linewidth}{
                %\centering
               \captionof{table}{DS2: \% of instances per solver label and category.}
      \label{tab:whereAreFragile}
                      \begin{tabular}{ccccc}
    \hline
        % & & Robust & \thead{Perturb\\-able}  & Fragile \\
           & & \rot{Robust} & \rot{Perturbable}  & \rot{Fragile}\\
         \hline
       \textbf{DS2} & BF & 29.0 & 10.5 &  10. 6\\
       &  FF  & 0.5 &  4.3 & 45.1 \\
       \hline
      \textbf{DS4 }& BF & 1.2 & 0.7 & 48.3 \\
       & FF & 0 & 0.8 & 49.0 \\
         \hline
    \end{tabular}
}
\end{table}

\section{Where are the fragile instances?}

Finally, we use a dimensionality-reduction technique to visualise  the instances in a 2d space to try to uncover any relationships between an instance (described by its data), the winning solver for the instance and the extent to which the original instances are: (a) \textit{fragile}: random search for a mask produces an adversarial sample; (b) \textit{perturbable}: evolution discovers at least one mask that creates an adversarial sample;
(c) \textit{robust }: an adversarial sample cannot be evolved.

\begin{figure}
    \centering
    \subfloat[Trained using solver labels \label{fig:UMAPDS2a}]
    {\includegraphics[width=0.49\textwidth]{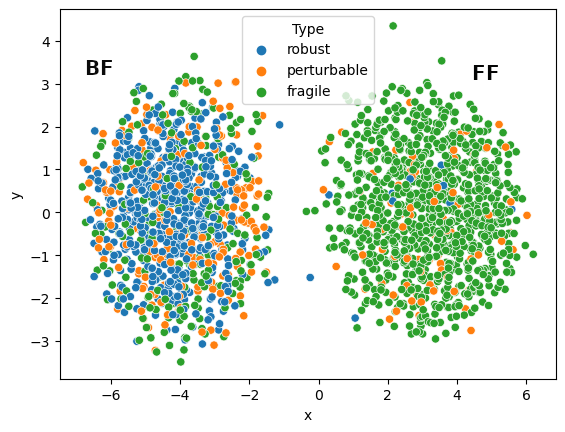}}
    \subfloat[Trained using `type' labels \label{fig:UMAPDS2b}]
    {\includegraphics[width=0.49\textwidth]{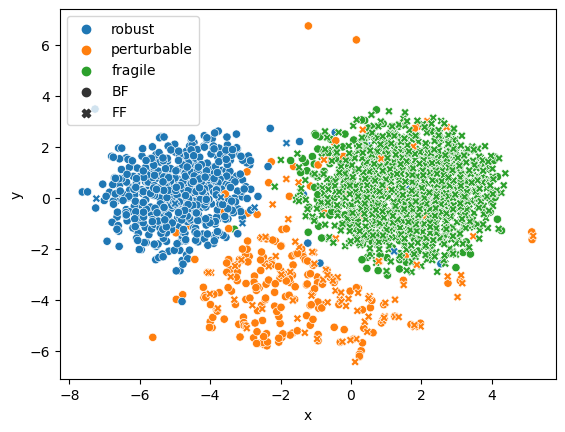}} 
    \caption{DS2(0.3), UMAP (supervised) trained with different labels.}
    \label{fig:UMAPDS2}
\end{figure}

In Figure~\ref{fig:UMAPDS2a}, we use supervised UMAP~\cite{mcinnes2018} to learn a projection that accounts for the solver label associated with the instance (BF/FF) --- this clearly separates the two classes. We then colour the instances according to the three categories above. 
It is immediately obvious that most of the instances in the FF cluster are fragile (see Table \ref{tab:whereAreFragile}); a small number are perturbable ($4.3\%$) and fewer than $1\%$ are robust. In contrast, most of the robust instances come from the BF class ($29\%$); in this class, the number of fragile and perturbable instances is approximately equal  $\approx 10\%$. In  Figure~\ref{fig:UMAPDS2b}, we again use supervised UMAP but this time train on the three category labels listed above. This clearly separates the three categories, again showing that the fragile instances mainly come from FF.  For DS4, the UMAP projection does not separate the three categories and is therefore not shown. Approximately $97\%$ of instances are fragile  and are uniformly distributed across the space, hence we omit this diagram. Statistics showing the percentage of instances per category for both datasets are given in Table~\ref{tab:whereAreFragile}.

\section{Conclusions and Future Work}

We investigated the robustness of a deep recurrent network used for algorithm-selection
to perturbations in instance data. This was inspired by the wealth of evidence from the image classification domain demonstrating that convolutional networks are particularly vulnerable to adversarial attacks. We proposed a method to evolve a mask that perturbs an instance such that the  modified instance is incorrectly classified by a DRN.
By restricting the level of perturbation allowed, the method ensures the evolved adversarial samples are similar to the original instances, therefore we expect the trained network should be capable of handling them.
However, using two datasets, we showed that instances can be categorised as \textit{fragile, perturbable} or \textit{robust} with respect to the trained models, 
and that adversarial samples can be evolved efficiently in between 1 and 112 generations, depending on the dataset and the initialisation method. Adversarial samples generated  are misclassified with a confidence $c$, where $0.5 < c \le 1.0$, i.e. $c$ ranges from very low to very high.

As well as bringing new insight into the robustness of the models, the approach sheds new light on which instances lie close to decision boundaries in the space, i.e. are easily perturbed.
We also found a subset of instances in which a perturbation causes the classifier to `flip' from classifying the instance correctly with very strong confidence to classifying incorrectly with equally strong confidence; every new perturbation can cause the instance to oscillate between these two states.  Further work is required to understand what characteristics of the instance data lead to this behaviour, and try to find features that correlate  with this. Another promising avenue of work would be to use an multi-objective algorithm to minimise the amount of per perturbation while maximising the probability of misclassification.
Finally, a  side-effect of the approach is that it generates a very large number of new instances. These instances are associated with a diverse range of classification probabilities and therefore represent a rich source of new training data for training better models in future.

\paragraph{Reproducibility:} code and data are available at~\cite{dataAdversarial}.

\begin{credits}
\subsubsection{\ackname} Emma Hart and Quentin Renau are supported by EPSRC EP/V026534/1

\clearpage
\subsubsection{\discintname}
The authors have no competing interests to declare that are
relevant to the content of this article. 
\end{credits}

\bibliographystyle{splncs04}
\bibliography{references}
\end{document}